\documentclass[runningheads,a4paper]{llncs}
\usepackage{dsfont}
\usepackage{amsmath}
\usepackage{amssymb}
\usepackage{graphicx}
\usepackage{subfigure}
\usepackage{wrapfig}
\usepackage{cite}
\usepackage{bm}
\usepackage{hyperref}
\usepackage{url}
\DeclareMathOperator*{\argmax}{arg\,max}
\DeclareMathOperator*{\argmin}{arg\,min}

\begin{document}

\mainmatter  

\title{Fiber Orientation Estimation Guided by a Deep Network}

\titlerunning{}

%
%
\author{Chuyang Ye$^{1}$ and Jerry L. Prince$^{2}$}
\authorrunning{Ye and Prince}

\institute{$^{1}$National Laboratory of Pattern Recognition \& Brainnetome Center, Institute of Automation, Chinese Academy of Sciences, Beijing, China\\
$^{2}$Department of Electrical and Computer Engineering, \\Johns Hopkins University, Baltimore, MD, USA}

%
%

\toctitle{}
\tocauthor{}
\maketitle

\begin{abstract}
\textit{Diffusion magnetic resonance imaging} (dMRI) is currently the only tool for noninvasively imaging the brain's white matter tracts. The fiber orientation (FO) is a key feature computed from dMRI for fiber tract reconstruction. Because the number of FOs in a voxel is usually small, dictionary-based sparse reconstruction has been used to estimate FOs with a relatively small number of diffusion gradients. 
However, accurate FO estimation in regions with complex FO configurations in the presence of noise can still be challenging.
In this work we explore the use of a deep network for FO estimation in a dictionary-based framework and propose an algorithm named \textit{Fiber Orientation Reconstruction guided by a Deep Network}~(FORDN). 
FORDN consists of two steps. First, we use a smaller dictionary encoding coarse basis FOs to represent the diffusion signals. To estimate the mixture fractions of the dictionary atoms (and thus coarse FOs), a deep network is designed specifically for solving the sparse reconstruction problem. Here, the smaller dictionary is used to reduce the computational cost of training.
Second, the coarse FOs inform the final FO estimation, where a larger dictionary encoding dense basis FOs is used and a weighted $\ell_{1}$-norm regularized least squares problem is solved to encourage FOs that are consistent with the network output. FORDN was evaluated and compared with state-of-the-art algorithms that estimate FOs using sparse reconstruction on simulated and real dMRI data, and the results demonstrate the benefit of using a deep network for FO estimation.

\keywords{diffusion MRI, fiber orientation estimation, deep network, sparse reconstruction}
\end{abstract}

\section{Introduction}
\label{sec:intro}

\textit{Diffusion magnetic resonance imaging} (dMRI) is currently the only tool that enables reconstruction of \textit{in vivo} white matter tracts~\cite{johansen}. By capturing the anisotropic water diffusion in tissue, dMRI infers information about fiber orientations (FOs), which are crucial features in white matter tract reconstruction~\cite{Basser}.

Various methods have been proposed to estimate FOs and handle situations where fibers cross. Examples include constrained spherical deconvolution~\cite{Tournier}, multi-tensor models~\cite{Ramirez,Landman,Aranda}, and ensemble average propagator methods~\cite{Merlet}. In particular, sparsity has shown efficacy in reliable FO estimation with a reduced number of gradient directions (and thus reduced imaging times)~\cite{Aranda}. The sparsity assumption is mostly combined with multi-tensor models~\cite{Ramirez,Landman,Aranda,Daducci,CMIG}, which leads to dictionary-based sparse reconstruction of FOs. However, accurate FO estimation in regions with complex FO configurations---e.g., multiple crossing fibers---in the presence of noise can still be challenging.

In this work, we explore the use of a deep network to improve dictionary-based sparse reconstruction of FOs. 
The deep network has drawn enormous attention in various computer vision tasks~\cite{Szegedy,Sun}; it is also shown to provide a promising approach to solving sparse reconstruction problems~\cite{Gregor,Xin,Sprechmann,Wang}. 
We model the diffusion signals using a dictionary, the atoms of which encode a set of basis FOs. Then, FO estimation can be formulated as a sparse reconstruction problem and we seek to solve it with the aid of a deep network. The proposed method is named \textit{Fiber Orientation Reconstruction guided by a Deep Network} (FORDN), which consists of two steps. First, a deep network that unfolds the conventional iterative estimation process is constructed and its weights are learned from synthesized training samples. To reduce the computational burden of training, this step involves only a smaller dictionary that encodes a coarse set of basis FOs, and thus gives approximate estimates of FOs. Second, the final sparse reconstruction of FOs is guided by the FOs produced by the deep network. A larger dictionary encoding dense basis FOs is used, and a weighted $\ell_{1}$-norm regularized least squares problem is solved to encourage FOs that are consistent with the network output. 
Experiments were performed on simulated and real brain dMRI data, where promising results were observed compared with competing FO estimation algorithms.

\section{Methods}
\label{sec:method}

\subsection{Background: FO Estimation by Sparse Reconstruction}

Diffusion signals can be modeled with a set of fixed prolate tensors, each representing a possible FO by its primary eigenvector (PEV)~\cite{Ramirez,Landman,Daducci}. Suppose the set of the basis tensors is $\{\mathbf{D}_{i}\}_{i=1}^{N}$ and their PEVs are $\{\bm{v}_{i}\}_{i=1}^{N}$, where $N$ is the number of the basis tensors. In practice, $N$ ranges from 100 to 300~\cite{Ramirez,Landman,CMIG}, and in this work we use $N=289$, which results from tessellating an octahedron. The eigenvalues of the basis tensors can be determined by examining the diffusion tensors in regions occupied by single tracts~\cite{Landman}.

For each diffusion gradient direction $\bm{g}_{k}$ ($k=1,\ldots,K$) associated with a $b$-value $b_{k}$, the diffusion weighted signal at each voxel can be represented as~\cite{Landman}
\begin{eqnarray}
S(\bm{g}_{k}) =
S(\bm{0})\sum\limits_{i=1}^{N}f_{i}e^{-b_{k}\bm{g}_{k}^{T}\mathbf{D}_{i}\bm{g}_{k}}
+ n(\bm{g}_{k}),
\label{eqn:dwi}
\end{eqnarray}
where $S(\bm{0})$ is the baseline signal without diffusion weighting, $f_{i}$ is the unknown nonnegative mixture fraction for $\mathbf{D}_{i}$ ($\sum_{i=1}^{N} f_{i} = 1$), and $n(\bm{g}_{k})$ represents image noise. By defining $y(\bm{g}_{k}) = S(\bm{g}_{k})/S(\bm{0})$ and $\eta(\bm{g}_{k}) = n(\bm{g}_{k})/S(\bm{0})$, we have
\begin{eqnarray}
\bm{y} = \mathbf{G}\bm{f} + \bm{\eta},
\label{equ:linear}
\end{eqnarray}
where $\bm{y}=(y({\bm{g}_1}),...,y({\bm{g}_K}))^{T}$, $\bm{f}=(f_{1},...,f_{N})^{T}$, $\bm{\eta}=(\eta({\bm{g}_1}),...,\eta({\bm{g}_K}))^{T}$, and $\mathbf{G}\in\mathbb{R}^{K\times N}$ is a dictionary matrix with $G_{ki}=e^{-b_{k}\bm{g}_{k}^{T}\mathbf{D}_{i}\bm{g}_{k}}$.

Because the number of FOs at a voxel is small compared with that of gradient directions, FOs can be estimated by solving a sparse reconstruction problem
\begin{eqnarray}
\hat{\bm{f}} = \argmin\limits_{\bm{f}\geq \bm{0},||\bm{f}||_{1}=1}||\mathbf{G}\bm{f}-\bm{y}||_{2}^{2} + \beta ||\bm{f}||_{0}\,.
\label{equ:sr}
\end{eqnarray}
To solve Eq.~(\ref{equ:sr}), the constraint of $||\bm{f}||_{1}=1$ is usually relaxed~\cite{Landman,Ramirez,Daducci}.
Then, in~\cite{Landman} and \cite{Ramirez} the problem is solved by approximating the $\ell_{0}$-norm with the $\ell_{1}$-norm; the authors of \cite{Daducci} solve the $\ell_{0}$-norm regularized least squares problem with iterative reweighted $\ell_{1}$-norm minimization~\cite{Candes}. The solution is finally normalized so that the mixture fractions sum to one and basis directions associated with mixture fractions larger than a threshold are set to be FOs~\cite{Landman}.

\subsection{FO Estimation Using a Deep Network}

Consider the general sparse reconstruction problem
\begin{eqnarray}
\hat{\bm{f}} = \argmin\limits_{\bm{f}}||\mathbf{G}\bm{f}-\bm{y}||_{2}^{2} + \beta ||\bm{f}||_{0}\,.  
\label{equ:sr2}
\end{eqnarray}
Using methods such as \textit{iterative hard thresholding}~(IHT)~\cite{Blumensath} or the \textit{iterative soft thresholding algorithm}~(ISTA)~\cite{Daubechies}, Eq.~(\ref{equ:sr2}) or its relaxed version with $\ell_{1}$-norm regularization can be solved by iteratively updating $\bm{f}$. At iteration $t+1$, 
\begin{eqnarray}
\bm{f}^{t+1} = h_{\lambda}(\mathbf{W}\bm{y}+\mathbf{S}\bm{f}^{t}),
\label{equ:iht}
\end{eqnarray}
where $\mathbf{W}=\mathbf{G}^{T}$, $\mathbf{S}=\mathbf{I}-\mathbf{G}^{T}\mathbf{G}$, and $h_{\lambda}(\cdot)$ is a thresholding operator with a parameter $\lambda\geq 0$. 
Motivated by this iterative process, previous works have explored the use of a deep network for solving sparse reconstruction problems.
By unfolding and truncating the process in Eq.~(\ref{equ:iht}), feed-forward deep network structures can be constructed for sparse reconstruction, where $\mathbf{W}$ and $\mathbf{S}$ are learned from training data instead of predetermined by $\mathbf{G}$~\cite{Gregor,Wang,Xin,Sprechmann}. 

\begin{figure}[!t]
  \centering
	\includegraphics[width=0.98\columnwidth]{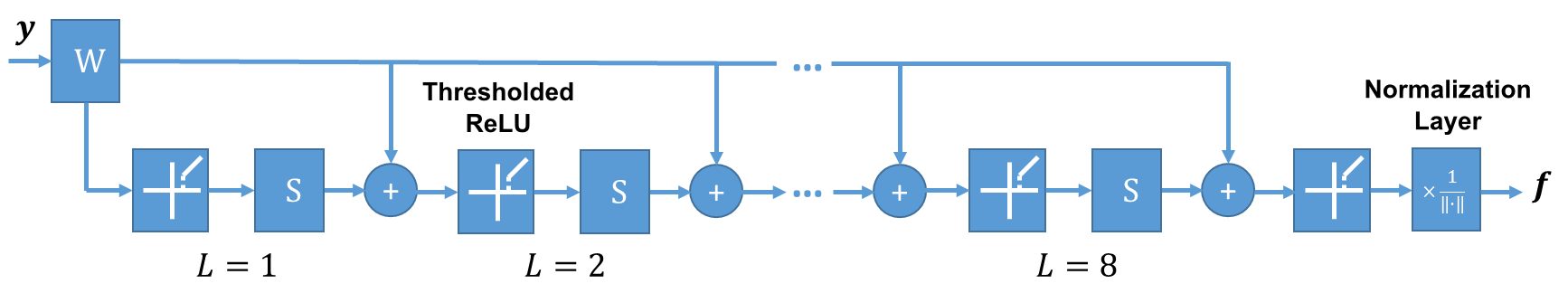}
\caption{The structure of the deep network used in this work for FO estimation.}
\label{fig:dn}
\end{figure}

In this work, to solve Eq.~(\ref{equ:sr}) we construct a deep network as shown in Fig.~\ref{fig:dn}.
The input is the diffusion signals $\bm{y}$ at a voxel and the output is the mixture fractions $\bm{f}$. The layers $L=1,2,\ldots,8$ correspond to the unfolded and truncated iterative process in Eq.~(\ref{equ:iht}) (assuming $\bm{f}^{0}=\bm{0}$), where $\mathbf{W}$ and $\mathbf{S}$ (shared among layers) are to be learned. A thresholded \textit{rectified linear unit} (ReLU)~\cite{Konda} is used in each of these layers
\begin{eqnarray}
[h_{\lambda}(\bm{a})]_{i} = 
\begin{cases}
0\quad &\mathrm{if}\quad a_{i} < \lambda\\
a_{i} \quad &\mathrm{if}\quad a_{i} \geq \lambda
\end{cases}
,
\label{equ:operator}
\end{eqnarray}
which corresponds to the thresholding operator in IHT~\cite{Blumensath}. We empirically set $\lambda=0.01$.
Note that because of the nonnegative constraint on $\bm{f}$ in Eq.~(\ref{equ:sr}), $[h_{\lambda}(\bm{a})]_{i}$ is always zero when $a_{i} < 0$.
A normalization layer is added before the output to enforce that the entries of $\bm{f}$ sum to one. To ensure numerical stability, we use $\bm{f}\leftarrow (\bm{f}+\tau\bm{1})/||\bm{f}+\tau\bm{1}||_{1}$ for the normalization, where $\tau=10^{-10}$. The network is implemented using the Keras library\footnote{\url{http://keras.io/}}. 
We use the mean squared error as the loss function and the Adam algorithm~\cite{Kingma} as the optimizer, where the learning rate is 0.001, the batch size is 64, and the number of epochs is 8.

Although it may seem ``wishful'' to expect a network with a small number of steps to beat the iterative process like IHT and ISTA, the authors of \cite{Gregor} argue that usually we do not seek to solve the problem for all possible inputs and only deal with a smaller problem where inputs resemble the training data. As well, the authors of \cite{Xin} demonstrate the benefit of using learned layer-wise fixed weights, where successful reconstruction can be achieved across a wider range of \textit{restricted isometry property}~(RIP) conditions than conventional methods such as ISTA and IHT.

Care must be taken in the construction of training data, because computing the sparse solution to Eq.~(\ref{equ:sr}) or (\ref{equ:sr2}) is an NP-hard problem. If the training data is generated by conventional algorithms, such as IHT and ISTA, then the network only learns a strategy that approximates these suboptimal solutions~\cite{Xin}. Thus, we adopt the strategy of synthesizing observations~\cite{Xin} according to given FO configurations. However, synthesis of diffusion signals for all combinations is prohibitive. For example, for the cases of three crossing fibers, the total number of FO combinations is ${N\choose 3}\approx 4\times10^{6}$ and each combination requires a sufficient number of training instances with noise sampling and different combinations of mixture fractions. This can be very computationally intensive for training the deep network. For example, training the deep network using the full set of basis directions failed on our Linux machine with 64GB memory due to insufficient memory. Therefore, motivated by the work of motion blur removal in~\cite{Sun}, we use a two-step strategy to estimate FOs. First, by using a smaller set of basis FOs, coarse FOs are estimated using the proposed deep network. Second, final FO estimation is guided by these coarse FOs by solving a weighted $\ell_{1}$-norm regularized least squares problem. Details of the two steps are described below.

\subsubsection{Coarse FO Estimation Using a Deep Network}

A smaller set of basis tensors $\{\tilde{\mathbf{D}}_{i'}\}_{i'=1}^{N'}$ ($N'=73$) and their PEVs $\{\tilde{\bm{v}}_{i'}\}_{i'=1}^{N'}$ are considered for coarse FO estimation using the deep network. 
As discussed and assumed in the literature, we consider cases with three or fewer FOs in synthesizing the training data~\cite{Daducci}. The cases of FO configurations can be given by applying an existing FO estimation method to the subject of interest. In this work we use CFARI~\cite{Landman} which estimates FOs using sparse reconstruction. Note that such a method does not need to provide accurate FO configurations at every voxel. Instead, it provides a good estimate of the set of FO configurations in the brain or a region of interest. 

\begin{figure}[!t]
  \centering
	\includegraphics[width=0.55\columnwidth]{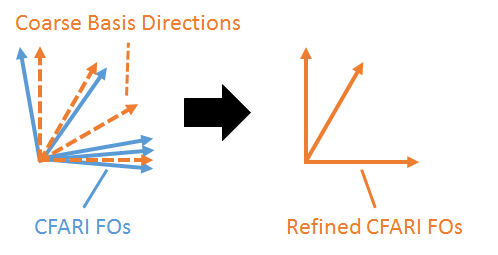}
\caption{An example of the CFARI FO refinement.}
\label{fig:refine}
\end{figure}

Because the original CFARI method can give multiple close FOs to represent a single FO that is not collinear with a basis direction, which unnecessarily increases the number of FOs, and that these FOs may not be collinear with the smaller set of basis directions considered in the deep network, we post-process the CFARI FOs $\{\bm{w}_{j}\}_{j=1}^{W}$ ($W$ is the number of CFARI FOs) associated with mixture fractions $\{h_{j}\}_{j=1}^{W}$ at each voxel (see Figure~\ref{fig:refine} for example). First, close FOs are refined so that only the peak directions are selected
\begin{eqnarray}
\resizebox{0.95\textwidth}{!} 
{$
\tilde{\mathcal{W}}=\{\tilde{\bm{w}}_{j'}^{}\}_{j'=1}^{\tilde{W}}=\{\bm{w}_{j}|\forall\:j'' \neq j \mbox{ and } \arccos(|\bm{w}_{j}\cdot\bm{w}_{j''}|)\leq
\frac{\pi}{180^\circ}\theta_{R}: h_{j}\geq h_{j''}\}.
\label{equ:refine} 
$}
\end{eqnarray}
Here, $\tilde{W}$ is the cardinality of $\tilde{\mathcal{W}}$ and we use $\theta_{R}=20^{\circ}$~\cite{FORNI}.
Then, these refined FOs $\tilde{\mathcal{W}}$ are mapped to their closest basis directions in $\{\tilde{\bm{v}}_{i'}\}_{i'=1}^{N'}$
\begin{eqnarray}
\tilde{\mathcal{V}}=\{\tilde{\bm{v}}_{n_{j'}} | n_{j'}=\argmax_{i'=1,2,\ldots,N'} |\tilde{\bm{v}}_{i'}\cdot\tilde{\bm{w}}_{j'}|\}.
\label{equ:map} 
\end{eqnarray}
$\tilde{\mathcal{V}}$ gives the FO configurations at each voxel represented by the coarse basis $\{\tilde{\bm{v}}_{i'}\}_{i'=1}^{N'}$ and each FO is only represented by one basis direction.

All post-processed FOs $\tilde{\mathcal{V}}$ in the brain or a brain region provide a set of training FO configurations. For each FO configuration with a single or multiple basis directions, diffusion signals were synthesized with a single-tensor or multi-tensor model using the corresponding basis tensors, respectively. For a single basis direction, its mixture fraction was set to one; for multiple basis directions, different combinations of their mixture fractions from 0.1 to 0.9 in increments of 0.1 were used for synthesis (note that they should sum to one). For example, the mixture fractions for three FOs can be (0.2,0.4,0.4). Rician noise was added to the synthesized signals, and the signal-to-noise ratio (SNR) can be obtained, for example, by placing bounding boxes in background and white matter areas~\cite{CMIG}. For each mixture fraction combination, 500 samples were generated for training.

Although the number of training samples is decreased by using a sparser set of basis directions, it may still consume a large amount of time and memory for training. To further reduce the computational cost of training, we parcellate the brain into different regions, each containing a smaller number of cases of FO configurations. This is achieved by registering the EVE template~\cite{Oishi} to the subject using the \textit{fractional anisotropy} (FA) map and the SyN algorithm~\cite{Avants}. A deep network is then constructed for each region using all the FO configurations in that region, and thus each network requires a much smaller number of training samples. This also has the benefit of letting each network deal with a smaller problem.

In the test phase, the trained networks estimate the mixture fractions in their corresponding parcellated brain regions. Like~\cite{Landman} and \cite{FORNI}, the basis directions with mixture fractions larger than a threshold of 0.1 are set to be the FOs and the FOs are also refined using Eq.~(\ref{equ:refine}). These refined FOs predicted by the deep networks are denoted by $\mathcal{U}=\{\bm{u}_{p}\}_{p=1}^{U}$ ($U$ is the cardinality of $\mathcal{U}$).

\subsubsection{FO Estimation Guided by the Deep Network}

The coarse FOs given by the deep networks provide only approximate FO estimates due to the low angular resolution of the coarse basis, however, they can guide the final sparse FO reconstruction that uses the larger set of basis directions. Specifically, at each voxel we solve the following weighted $\ell_{1}$-norm regularized least squares problem~\cite{CMIG} that allows incorporation of prior knowledge of FOs,
\begin{eqnarray}
\hat{\bm{f}} = \argmin\limits_{\bm{f}\geq \bm{0}}||\mathbf{G}\bm{f}-\bm{y}||_{2}^{2} + \beta ||\mathbf{C}\bm{f}||_{1}\,.  
\label{equ:wsr}
\end{eqnarray}
Here, $\mathbf{C}$ is a diagonal matrix encoding the guiding FOs predicted by the deep network, and basis directions closer to the guiding FOs are encouraged. We use the design given by~\cite{FORNI}, where the diagonal weights are specified as
\begin{equation}
C_{i} = \frac{1 - \alpha\max\limits_{p=1,\ldots,U} |\bm{v}_{i}\cdot
  \bm{u}_{p}|}{\min\limits_{q=1,\ldots,N}\left(1 -
    \alpha\max\limits_{p=1,\ldots,U}|\bm{v}_{q}\cdot \bm{u}_{p}|\right)}, 
\quad i=1,\ldots, N \,.  
\label{eq:diagonalweights}
\end{equation}
When the basis direction $\bm{v}_{i}$ is close to the guiding FOs, its weight $C_{i}$ is small; thus, $f_{i}$ is encouraged to be large and $\bm{v}_{i}$ is encouraged to appear. Eq.~(\ref{equ:wsr}) can be solved using the strategy given by~\cite{CMIG}. We set $\alpha=0.8$ as in~\cite{FORNI}, and selected $\beta=0.25$ because the number of diffusion gradients used in this work is about half of that used in~\cite{FORNI}. The mixture fractions are normalized so that they sum to one, and the FOs are determined as the basis directions associated with mixture fractions larger than 0.1~\cite{Landman,CMIG} and refined using Eq.~(\ref{equ:refine}).

\section{Results}
\label{sec:exp}

\subsection{3D Digital Crossing Phantom}
\label{sec:exp_phantom}

\begin{figure}[!t]
  \centering
	\includegraphics[width=0.85\columnwidth]{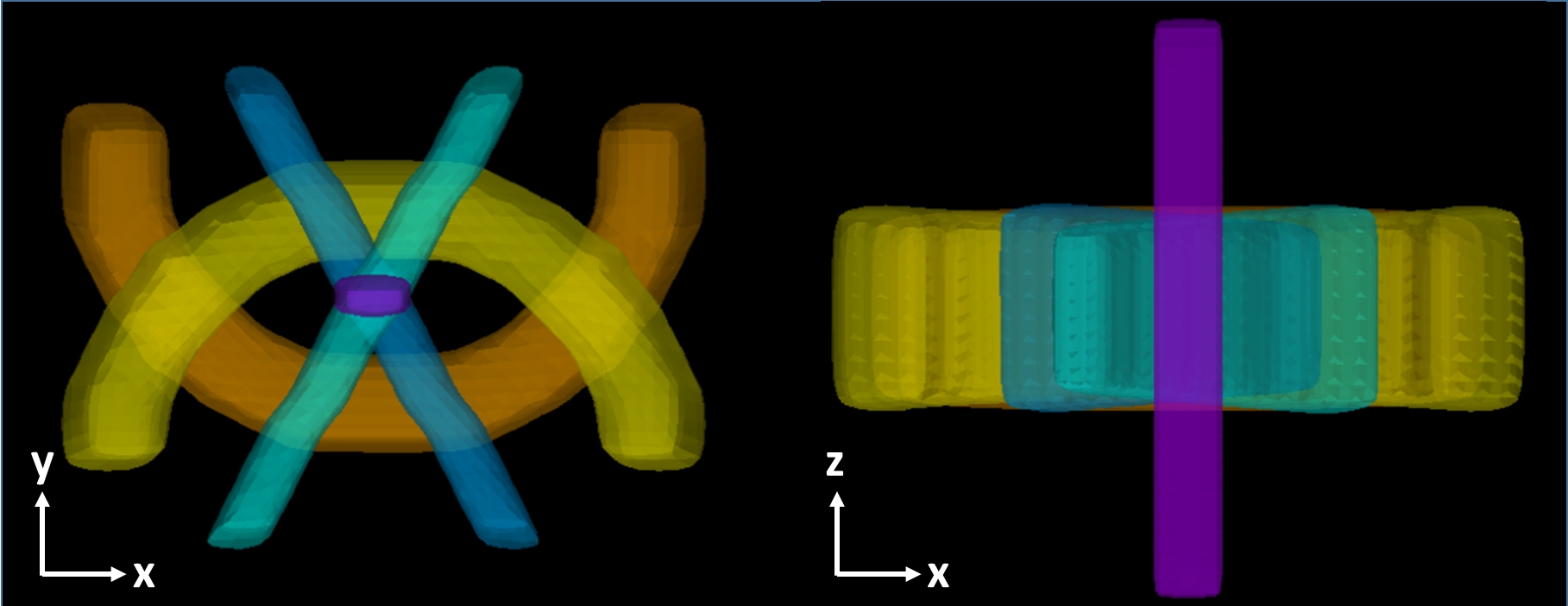}
\caption{A 3D phantom simulating five fiber tracts.}
\label{fig:phantom}
\end{figure}

A 3D digital crossing phantom was created to simulate five tracts that cross in pairs at six locations and triplets at one location (see Figure~\ref{fig:phantom}), where the tract geometries and diffusion parameters in~\cite{FORNI} were used. Thirty gradient directions were applied with $b=1000~\mathrm{s}/\mathrm{mm}^2$. Rician noise ($\mathrm{SNR}=20$ on the b0 image) was added to the \textit{diffusion weighted images} (DWIs).

\begin{figure}[!t]
  \centering
	\includegraphics[width=0.55\columnwidth]{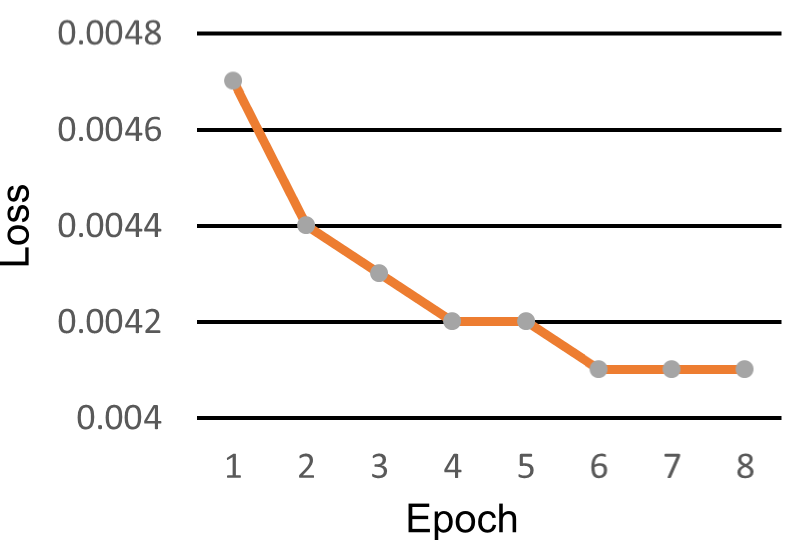}
\caption{Values of the loss function after each epoch in the training process.}
\label{fig:loss}
\end{figure}

FORDN was applied to the phantom data. Note that here we did not parcellate the phantom into different subregions because the number of FO configurations is similar to that in a typical brain region parcellated by the EVE atlas. The training process took about 10 minutes on a 16-core Linux machine and the training loss after each epoch is plotted in Figure~\ref{fig:loss}. With the selected parameters, the training loss becomes stable after 8 epochs. 

\begin{figure}[!t]
  \centering
	\includegraphics[width=0.9\columnwidth]{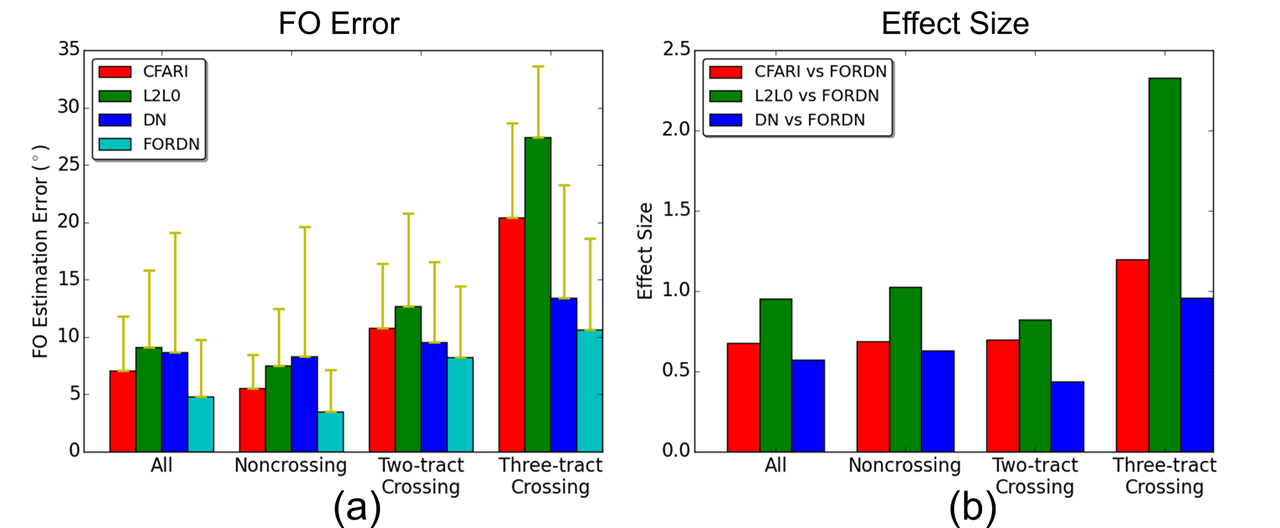}
\caption{(a) Means and standard deviations of FO errors. (b) Effects sizes for the comparison between FORDN and other methods.}
\label{fig:error}
\end{figure}

We then quantitatively evaluated the accuracy of FORDN using the error measure proposed in~\cite{FORNI}, and compared it with two competing methods, CFARI~\cite{Landman} and L2L0~\cite{Daducci}, because like FORDN they achieve voxelwise FO estimation using sparse reconstruction. We also compared the final FORDN results with the intermediate output from the deep network (DN). 
The errors in the entire phantom and in each region containing noncrossing, two crossing, or three crossing tracts are shown in~Figure~\ref{fig:error}(a). In all cases, FORDN achieves more accurate FO reconstruction. In addition, the intermediate DN results already improves FO estimation in regions with crossing tracts compared with CFARI and L2L0.

The FO errors in Fig.~\ref{fig:error}(a) were also compared between FORDN and CFARI, L2L0, and DN using a paired Student's $t$-test. 
In all cases, the FORDN errors are significantly smaller ($p<0.001$), and the effect sizes (Cohen's $d$) are shown in Figure~\ref{fig:error}(b). The effect sizes are larger in regions with three crossing tracts, indicating greater improvement in regions with more complex fiber structures.

\subsection{Brain dMRI One}
\label{sec:brain1}

Next, FORDN was evaluated on a brain dMRI scan. DWIs were acquired on a 3T MR scanner (Magnetom Trio, Siemens, Erlangen, Germany), which include thirty gradient directions with $b=1000~\mathrm{s}/\mathrm{mm}^2$. The resolution is 2.7 mm isotropic. The SNR is approximately 20 on the b0 image.

\begin{figure}[!t]
  \centering
	\includegraphics[width=0.85\columnwidth]{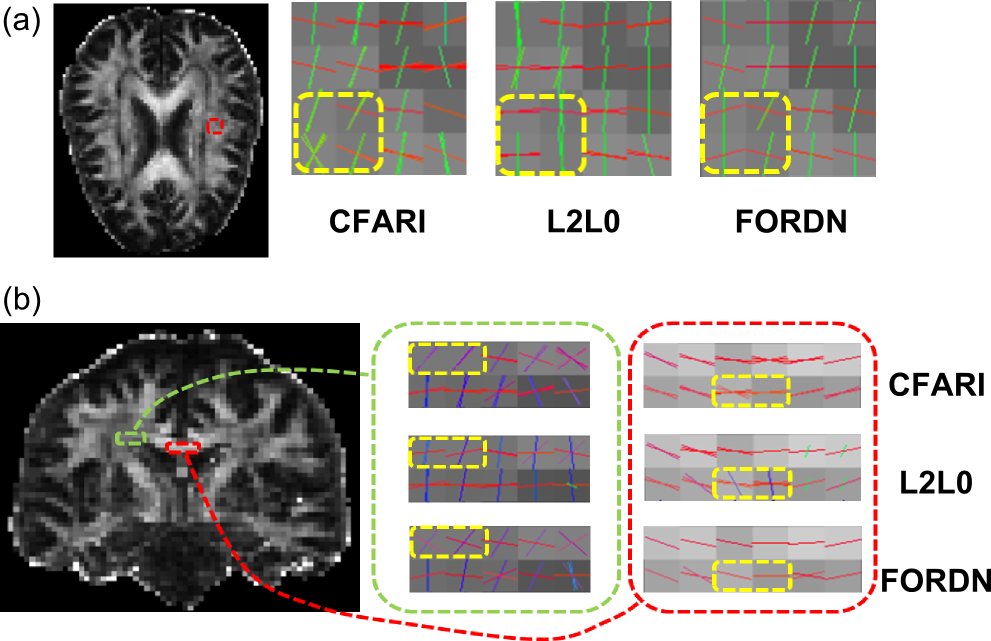}
\caption{FO estimation results of brain dMRI one overlaid on the FA map: (a) an axial view of the crossing of the CC and SLF and (b) a coronal view of the noncrossing CC and the crossing of the CC and CST. Note the highlighted region for comparison.}
\label{fig:brain1}
\end{figure}

We selected regions (see Figure~\ref{fig:brain1}) containing the \textit{corpus callosum} (CC) for evaluation. The FOs reconstructed in the selected regions are shown and compared with CFARI and L2L0 in Figure~\ref{fig:brain1}. In Figure~\ref{fig:brain1}(a), FOs in the region where the CC and the \textit{superior longitudinal fasciculus} (SLF) cross are displayed, and FORDN better identifies the crossing (note the highlighted region for comparison). In Figure~\ref{fig:brain1}(b), a region (red box) containing the noncrossing CC and a region (green box) where the CC and the \textit{corticospinal tract} (CST) cross are selected. In the noncrossing CC, CFARI and FORDN do not generate the false FOs in the L2L0 result (see the vertical FOs in the highlighted region); in the crossing of the CC and CST, L2L0 and FORDN better identify the crossing FOs than CFARI (see the highlighted region).

\subsection{Brain dMRI Two}
\label{sec:brain2}

FORDN was also applied to a dMRI scan of a random subject from the Kirby21 dataset~\cite{Kirby}. DWIs were acquired on a 3T MR scanner (Achieva, Philips, Best, Netherlands). Thirty-two gradient directions ($b=700~\mathrm{s}/\mathrm{mm}^2$) were used. The in-plane resolution is 2.2~mm isotropic and was upsampled by the scanner to 0.828~mm isotropic. The slice thickness is 2.2~mm. We resampled the DWIs so that the resolution is 2.2 mm isotropic.
The SNR is about 22 on the b0 image.

\begin{figure}[!t]
  \centering
	\includegraphics[width=0.85\columnwidth]{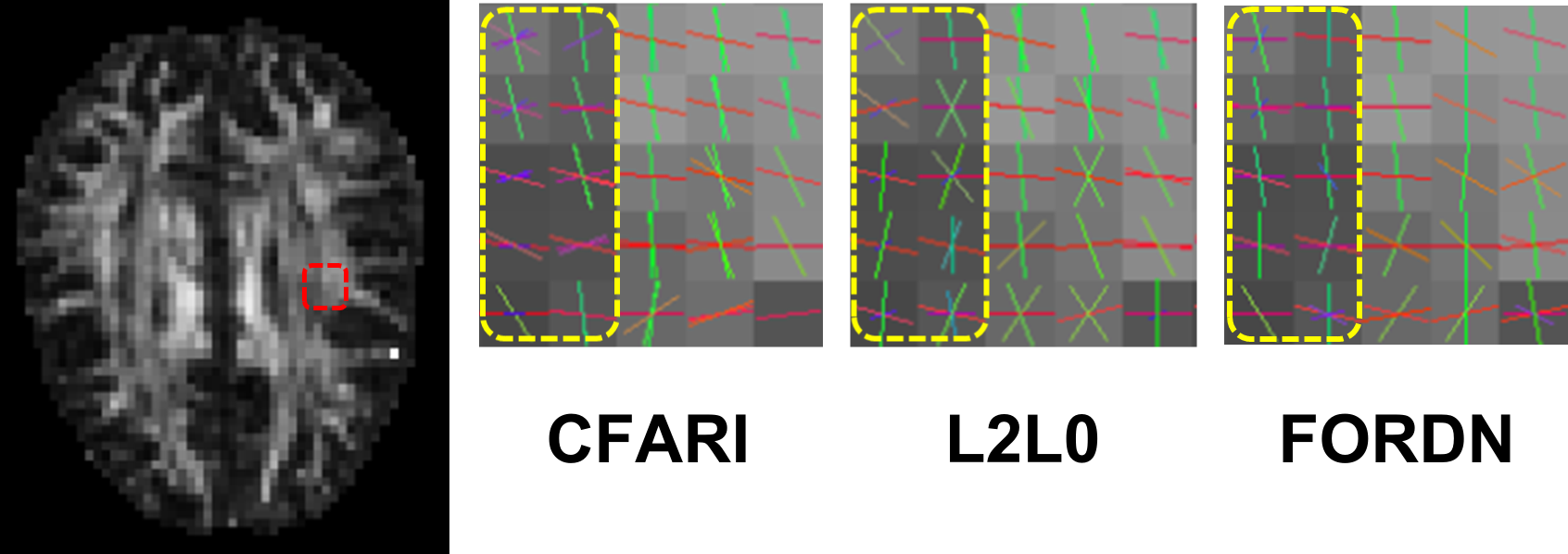}
\caption{FO estimation results of brain dMRI two overlaid on the FA map: an axial view of the crossing of the CC and SLF. Note the highlighted region for comparison.}
\label{fig:brain2}
\end{figure}

\begin{figure}[!t]
  \centering
	\includegraphics[width=0.95\columnwidth]{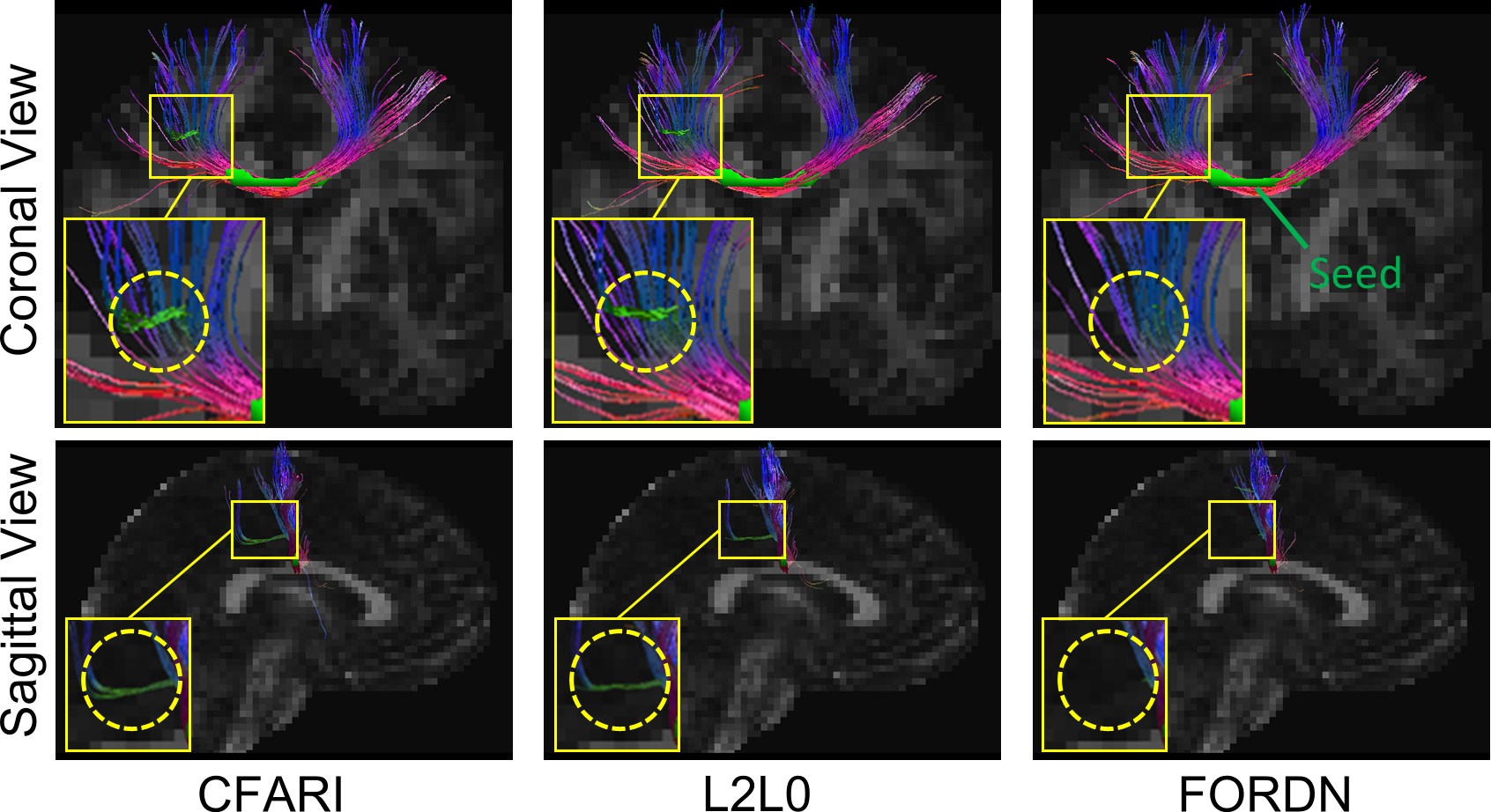}
\caption{Fiber tracking results seeded in the CC. Note the zoomed region for comparison.}
\label{fig:ft}
\end{figure}

FOs in a region where the CC and SLF cross are shown in Figure~\ref{fig:brain2} and compared with CFARI and L2L0. FORDN better reconstructs the transverse CC FOs and the anterior--posterior SLF FOs than CFARI and L2L0 (see the highlighted region for example). Fiber tracking was then performed using the strategy in~\cite{Yeh}, where seeds were placed in the noncrossing CC (see Figure~\ref{fig:ft}). The FA threshold is 0.2, the turning angle threshold is $45^{\circ}$, and the step size is 1~mm. The results are shown in Figure~\ref{fig:ft}, and each segment is color-coded using the standard color scheme~\cite{Pajevic}. FORDN FOs do not produce the false (green) streamlines going in the anterior--posterior direction as in the CFARI and L2L0 results (see the zoomed region). Note that the streamlines tracked by FORDN propagate through multiple regions parcellated by the EVE atlas, which indicates that the consistency of the fiber streamlines is preserved although each region is associated with a different deep network.
\section{Discussion}
\label{sec:discussion}

The parameters selected in this work produced reasonable results, including convergence of training loss and desirable FOs. A thorough investigation of the impact of these parameters may be performed in the future.
It is also possible to learn the parameter $\lambda$ in the network. For example, in~\cite{Wang} a division layer and a multiplication layer with the parameter $\lambda$ are added before and after the activation unit (the Thresholded ReLU in Fig.~\ref{fig:dn}), respectively, and the parameter in the activation unit is set to one. Thus, $\lambda$ becomes a parameter that can be optimized in the training. Such a strategy can be explored in future work. 

Previous works~\cite{CMIG,FORNI} have used weighted $\ell_{1}$-norm regularization to guide FO estimation. In~\cite{CMIG}, tongue muscle FO estimation is informed by the geometry of the muscle fiber tracts, which is possible due to their simple shapes; however, in the brain the fiber tracts are much more complex and this strategy is not applicable in general. 
In~\cite{FORNI}, the guiding information is extracted from neighbor FO information, but the FOs in the neighbors are also to be estimated. 
Our use of a deep network can be interpreted as an approach to prior knowledge incorporation for sparse FO reconstruction in the brain.
It would also be interesting to incorporate spatial consistency of FOs in the network for improved FO estimation, where the network could take image patches as input and the interaction between neighbor voxels must also be designed in the network.

We have assumed a fixed dictionary for FO estimation, yet variability of the dictionary can exist at different locations. \cite{Aranda} and \cite{Yap} have used different strategies to account for this issue. It is possible to extend our deep network to include multiple fiber response functions in the dictionary like~\cite{Yap}.

\section{Conclusion}
\label{sec:conclusion}

We have proposed an algorithm of FO estimation guided by a deep network. The diffusion signals are modeled in a dictionary-based framework. A deep network designed for sparse reconstruction provides coarse FO estimation using a smaller set of the dictionary atoms, which then informs the final FO estimation using weighted $\ell_{1}$-norm regularization. Results on simulated and brain dMRI data have demonstrated promising results compared with the competing methods.

%

\bibliographystyle{splncs03}
\bibliography{refs}
\end{document}